\documentclass{bmvc2k}

\title{Maturity-Aware Active Learning for Semantic Segmentation with Hierarchically-Adaptive Sample Assessment }

\addauthor{Amirsaeed Yazdani}{yazdaniamir38@yahoo.com}{1}
\addauthor{Xuelu Li}{xueluli122@gmail.com}{1}
\addauthor{Vishal Monga}{vum4@psu.edu}{1}

\addinstitution{
Department of Electrical Engineering\\ 
The Pennsylvania State University\\
University Park, USA
}

\runninghead{Yazdani \emph{\MakeLowercase{et al}.}}{AL with Hierarchically-Adaptive Sample Assessment}

\usepackage{mathtools}
\usepackage{wrapfig}
\DeclarePairedDelimiter\ceil{\lceil}{\rceil}
\begin{document}
\nocite{CCT}
\maketitle
\begin{abstract}
Active Learning (AL) for semantic segmentation is challenging due to heavy class imbalance and different ways of defining ``sample'' (pixels, areas, etc.), leaving the interpretation of the data distribution ambiguous. We propose ``Maturity-Aware Distribution
Breakdown-based Active Learning'' (MADBAL), an AL method that benefits from a hierarchical approach to define a multiview data distribution, which takes into account the different "sample" definitions jointly, hence able to select the most impactful segmentation pixels with comprehensive understanding. MADBAL also features a novel uncertainty formulation, where AL supporting modules are included to sense the features' maturity whose weighted influence continuously contributes to the uncertainty detection. In this way, MADBAL makes significant performance leaps even in the early AL stage, hence reducing the training burden significantly. It outperforms state-of-the-art methods on Cityscapes and PASCAL VOC datasets as verified in our extensive experiments.
\end{abstract}

%------------------------------------------------------------------------- 
\section{Introduction}
\label{sec:intro}
Neural networks in the past decade have been dominant solutions for a wide majority of computer vision problems \cite{relighting,PA-TMI,classification,seg1,xuelubmvc}; however, these solutions often suffer from being data-eccentric, which means a burden in both the data collection and annotation. While this burden exists for almost every computer vision task, it becomes more costly and laborious for tasks that need fine-grained annotations, such as image segmentation.

Active learning (AL) methods have been proposed to overcome this bottleneck by incrementally selecting the samples for improving the performance of the current model, which has been trained on a limited training set. AL offers some criteria based on which the samples in an unlabeled pool are assessed, ranked, selected, and then added to the current training pool. The majority of existing AL methods \cite{Classal1,detal1,AL_uncertainty,AL_diversity} rely on a combination of the criteria such as model uncertainty and the diversity in the labeled pool. Although these criteria seem intuitive and are well-defined for tasks such as classification, for segmentation their definition becomes ambiguous, hence challenging to quantify. This is due to the fact that in segmentation, ``sample'' does not have a concrete definition based on which the formulation can uniquely rank the samples like classification. In segmentation, a ``sample'' can be perceived either as a pixel, a patch, or an image. Moreover, the informativeness of the samples for a specific definition is correlated to the information of surrounding samples. This leaves the existing segmentation AL methods \cite{SAL,rectAL,pixelpick}, focusing on only one interpretation of samples, with a narrow insight into the data distribution.    

\begin{figure}[!ht]
  \centering
  \includegraphics[width=\linewidth]{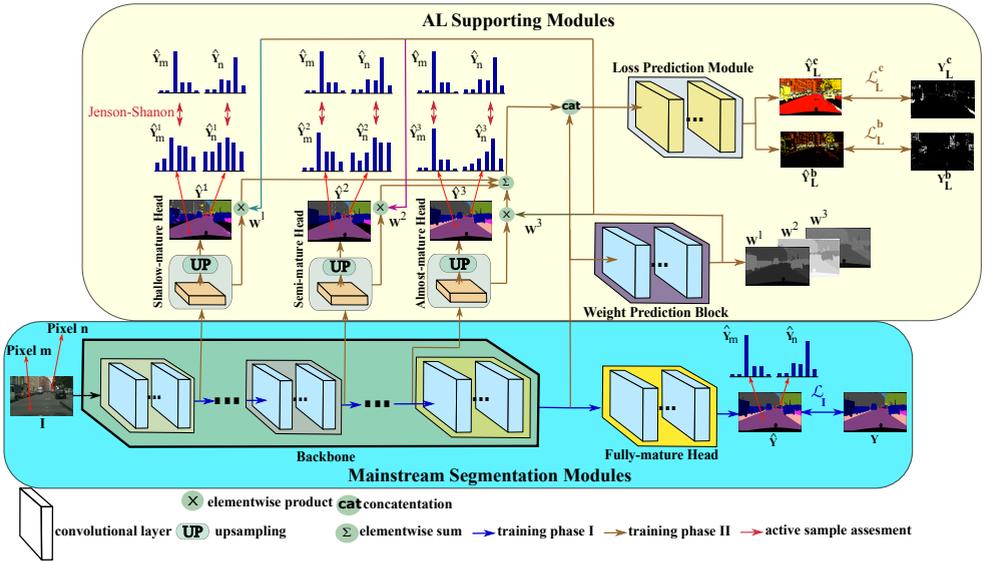}
  \caption{The proposed backbone-agnostic architecture. We use pixels m and n as examples of easy and challenging samples to understand sample assessment mechanism. For the easy sample, the predicted probability distributions ($\hat{Y}_{m}^{i}$) show high similarity to the final distribution ($\hat{Y}_{m}$) (measured by Jenson-Shannon divergence). For the challenging sample, the predicted probabilities ($\hat{Y}_{n}^{i}$) show a confusing trend at shallow stages, taking longer to show a consistent trend and high similarity to the final distribution ($\hat{Y}_{n}$). For the details of the modules and their training schemes refer to \ref{modules}.}
  \label{fig:arch}
\end{figure}

\begin{wrapfigure}[20]{r}{0.7\linewidth}
  \begin{center}
    \includegraphics[width=1\linewidth]{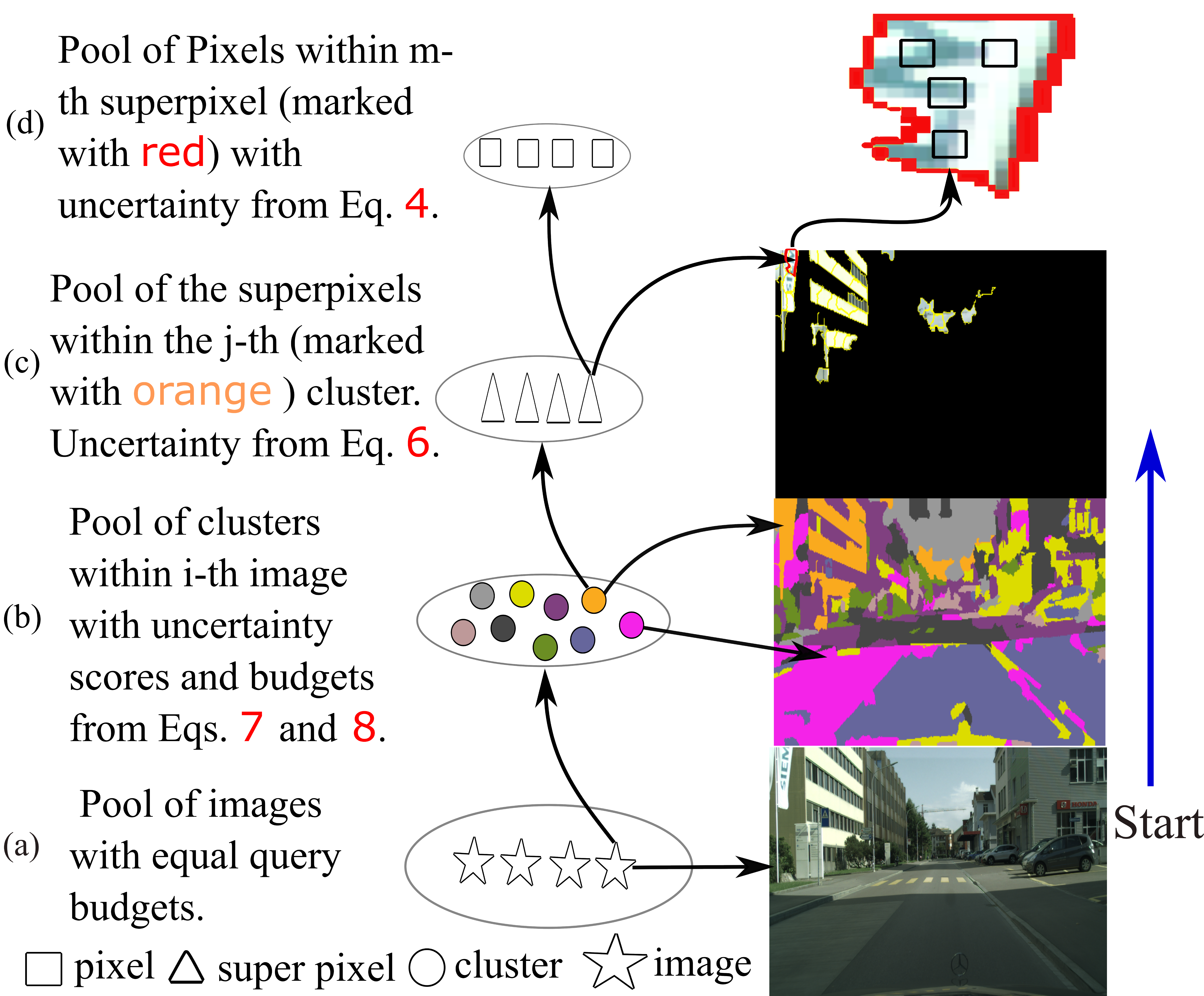}
  \end{center}
  \vspace{-10pt}
  \caption{Our hierarchical distribution breakdown approach.}
  
      \label{fig:pyramid}
      \vspace{-12pt}
\end{wrapfigure}

 In this work, we propose a systematic and inclusive AL strategy as a natural evolution of existing works with different sample considerations,\ tied with backbone-agnostic, AL supporting network components. Specifically, our key contributions are:  
\begin{itemize}
    \item \textbf{Distribution Breakdown}: We propose a \textbf{hierarchical approach} to estimate the data distribution based on different definitions of ``sample,'' which allows for a multilevel assessment of the data. We traverse this hierarchy level by level, while at each level breaking down the distribution of the data according to the corresponding ``sample'' definition (see Fig. \ref{fig:pyramid}). This means that the representativeness of the selected data is checked across multiple views making the training set as insightful as possible.
    \item\textbf{Maturity-Awareness}: We propose a set of \textbf{backbone-agnostic, AL supporting modules} associated with \textbf{carefully devised uncertainty terms} which together are capable of detecting the most impactful samples for network performance improvement. AL supporting modules help monitor the flow of information through different layers with different features' maturity level (see Fig. \ref{fig:arch}). This flow is interpreted via our proposed uncertainty formulation which evaluates the model maturity for different samples.
\item Integration of the aforementioned algorithmic pieces results in a model referred to as ``\textbf{M}aturity-\textbf{A}ware \textbf{D}istribution \textbf{B}reakdown-based \textbf{A}ctive \textbf{L}earning'' (\textbf{MADBAL}). We evaluate the performance of our model on Cityscapes \cite{cityscapes} and Pascal VOC 2012 \cite{VOC} datasets and prove that not only does MADBAL outperform state of the art w.r.t different metrics, but also exhibits immediate performance leaps unlike state of the art where the improvements are more gradual. This makes MADBAL a preferred AL solution for reducing the training burden overhead from two  standpoints: 1) \textbf{Lower number of AL steps} needed for achieving acceptable results. 2) \textbf{No requirement for a rich, carefully-selected initial labeled pool}.
\end{itemize}

\section{Related Works}
 \textbf{Selection criteria} in the AL literature consist mainly of two types of nature: \textbf{uncertainty} and \textbf{diversity}. Uncertainty-based criteria \cite{AL_uncertainty} focus on how certain the model is in its prediction for a possible candidate and select ones with more uncertainty. They can be mathematically formulated in a variety of ways such as posterior probability of the predicted class \cite{AL_posterior} or the margin between the posterior probabilities of the predicted class and the the class that received the second highest predicted probability \cite{AL_margin}. This complements diversity-based criteria \cite{AL_diversity} with the main objective to help the training set maintain a representation as close as possible to the whole distribution of the data.\ This would lead to detection and addition of the samples so that the distance between the training set and the unlabeled pool is minimized (Core-set AL) \cite{coreset}, or the most representative subset of the unlabeled pool is constructed \cite{AL_opt}. AL methods consider one or a hybrid combination of these two criteria for their selection. For example, BALD \cite{Bald} uses a Bayesian framework to select samples based on the uncertainty of sampled networks. Later, BatchBALD \cite{BatchBald} was proposed as a modification of BALD to take the diversity into consideration. Besides these criteria, \textbf{expected model change} is used in a few works \cite{EMC1,EMC2,EMC3,EMC4} as a criterion to select samples that cause the greatest change in the current model or its output. For example, Freytag \emph{et al.} \cite{EMC4} use the current model to predict the output changes, while Settles \emph{et al}. \cite{EMC1} rely on the predicted gradient length to select samples.   

\textbf{AL methods for semantic segmentation} are categorized into image-based and region-based methods:
\textbf{Image-based methods} \cite{hueristic,VAAL,deal,cost} are often faster with lower computational complexity owing to their definition of ``sample,'' which gives them smaller search space at the expense of adding redundant classes at every AL step. This in turn leads to less budget-efficiency of these approaches. As an image-based method for medical image segmentation, Yang \emph{et al}. \cite{hueristic} propose a CNN architecture and a heuristic method to find the most representative samples among top k with highest uncertainty. Within the same domain, \cite{cost} leverages drop-out to represent the Monte Carlo sampling at test time for melanoma segmentation. \cite{VAAL} leverages the min-max game between the adversarial network and the variational autoencoder (VAE) to discriminate between challenging and easy samples based on the features in the latent space. Inspired by the work of Yoo \emph{et al}. \cite{learningloss} for dedicating network components for loss prediction, Xie \emph{et al}. \cite{deal} develop a difficulty-aware network to generate difficulty heatmaps using the missclassified/correctly classified pixels in the labeled pool. \textbf{Region-based methods} \cite{pixelpick,SAL,Viewal,CPRAL,cereals}, unlike image-based methods, show higher performance with significantly lower budget as they are able to select only the regions with the most helpful classes for annotation, hence no need for annotating useless regions. This has led to emergence of more region-based methods recently. CEREALS \cite{cereals} estimates the cost of annotating regions and finds a trade-off between the informativeness and annotation cost of the candidates. Golestaneh \emph{et al}. \cite{equal} utilize the fact that the most uncertain regions show high uncertainty under equivariant transformations. Recently, Cai \emph{et al}. \cite{SAL} was one of the pioneers in estimating the data distribution by using the trained model at the current step to find the dominant labels across superpixels and select the most uncertain superpixels whose dominant labels belong to less frequent classes, which inspires our uncertainty formulation at the superpixel level; nevertheless we extend \cite{SAL} by introducing AL supporting modules at other levels. Focusing on pixels, PixelPick \cite{pixelpick} in each round of AL selects an equal number of pixels with highest uncertainty from each image. Recently, \cite{CPRAL} deploys a regional Gaussian attention module to select regions and leverages contextual guidance to extend the regional annotations to unlabeled regions, while borrowing the idea of the loss prediction module from \cite{deal}. The proved benefits of loss prediction module in \cite{learningloss,deal,CPRAL} motivates us in including it in our AL supporting modules; however, as it will be elaborated, ours benefits from a more effective training protocol (separate training phases), allocating boundary-aware output channels, and more effective ground truth formulation.       
\vspace{-10px}
\section{Methods}
 Our method includes two main components: 1) AL supporting modules, components of which reflect the \textbf{information flow needed for maturity-awareness}, and 2) selection strategy, which reflects our \textbf{hierarchical distribution breakdown scheme} integrated with our \textbf{custom uncertainty formulation}.
\raggedbottom
% \subsection{Network architecture}
% The proposed network features modules and layers which can be categorized into two different sectors: 1) mainstream segmentation modules and 2) AL supporting modules. We would need different training phases for these modules which will be explained along with each sector. 
% \raggedbottom
% \subsubsection{Mainstream Segmentation Modules}
% The mainstream segmentation modules are essential components for learning the main task and include backbone and fully-mature head (connected with blue arrows in Fig. \ref{fig:arch}). Through training phase I, the loss function ($ L_{\textit{phase I}}$) for training is the weighted cross-entropy loss. This way, we ensure the modules dedicated to the main task are trained based on their primary purpose without any additional auxiliary concentration which might distract them from their optimal performance.
% \begin{equation}
%     \mathcal{L}_{\textit{phase I}}(\theta_{\textit{backbone}},\theta_{\textit{seg head}})=\frac{1}{N}\sum_{i=1}^{N}-W_{Y_{i}}log(\hat{Y}_{i,Y_{i}})
% \end{equation}
% Where N, $Y_{i}$, $\hat{Y}_{i,c}$, and $W_{c}$  denote the total number of labelled pixels within a batch, the ground truth class label for pixel i, the output logit for class c for pixel i, and the weight for class c, respectively. $\theta_{\textit{backbone}}$ and $\theta_{\textit{seg head}}$ are the parameters of the backbone and segmentation head, respectively.  
\subsection{AL Supporting Modules}
\label{modules}
Our network consists of conventional modules essential for carrying out the main segmentation task (Mainstream Segmentation modules -- see Fig. \ref{fig:arch}) which makes it backbone-agnostic. These modules are trained through a preliminary phase of training (training phase I) with cross-entropy (CE) loss. Once these modules are trained, the training of AL supporting modules, whose purpose is critical at the time of sample selection, starts based on their designated goal.\\ 
\indent\textbf{Varied-Maturity Heads} besides the main segmentation head (Fully-mature head -- see Fig. \ref{fig:arch}), include three heads with access to different depths of the backbone layers. Indeed, as the depth increases, the maturity of features provided to these heads increases. Starting from the shallowest, we denote them with \textbf{Shallow-mature Head}, \textbf{Semi-mature Head}, and \textbf{Almost-mature Head}. Each head in training phase II is trained for the segmentation task and assigned a loss term ($\underset{k \in \{1,2,3\} }{\operatorname{\mathcal{L}^{k}_{\textit{seg}}}}$) which is CE loss defined on segmentation outputs.\\ 
\indent \textbf{Loss Prediction Module} is in charge of predicting the probability of each pixel's error-proneness for the segmentation task. This module makes use of a different version of the feature maps provided to the varied-maturity heads and the endpoint features of the backbone (see Fig. \ref{fig:arch}). These features are weighted by the weight maps provided by the \textbf{Weight Prediction Block} to help the model reweight the features based on their importance. We define the ground truth for the loss prediction task by considering the class-specific average loss across the labeled pool as a threshold for determining the loss labels:
\begin{equation}
\label{eq:loss_label}
    Y^{i}_{L}= 
\begin{cases}
    1,& \text{if } L_{\textit{I}}^{i}\geq \tau_{Y_{i}}\\
    0,              & \text{otherwise}
\end{cases}
\end{equation}
Where $ Y^{i}_{loss}$, $ \tau_{c}$, and $L_{\textit{phase I}}^{i}$ are the loss label for pixel $i$, the phase I mean loss (CE) across all the pixels belonging to class $c$ in the labeled pool, and phase I loss for pixel $i$, respectively. Since each pixel is labeled based on how it compares to other members of its class, the model acquires a more insightful loss prediction capability specialized for each class. Moreover, as the model is already trained for the main task, the ground truth does not change during the training of this module unlike existing works \cite{deal,CPRAL}. Next, we follow the training for the loss prediction task separately for the pixels lying on the boundary and center regions via assigning two output channels. This aids the module for a better focus on different levels of error as it is known that the segmentation error is generally higher on the boundary of the objects \cite{boundary1,boundary2}. Thus, the loss for loss prediction module would be:
\setlength{\abovedisplayskip}{0pt}
\setlength{\belowdisplayskip}{3pt}
\begin{multline}
\label{eq:loss_pred}
      \mathcal{L}_{\textit{L}}^{m\in \{\textit{c, b}\}   }(\theta_{\textit{ seg heads}},\theta_{\textit{L}},\theta_{\textit{W}})=\\
      \frac{1}{|\{x:x \in m\}|}\sum_{\{i:x_{i} \in m\}}\Bigl(-Y_{i}^{L}log(\sigma(\hat{Y}^{i}_{L-m}))-(1-Y_{i}^{L})log(1-\sigma(\hat{Y}^{i}_{L-m}))\Bigr)
\end{multline}
Where $\theta_{\textit{seg heads}}$, $\theta_{\textit{L}}$, and $\theta_{\textit{W}}$ denote the parameters of varied-maturity segmentation heads, loss prediction module, and weight prediction block, respectively. Additionally, $|.|$, $\sigma(.)$, and $\hat{Y}^{i}_{L-m}$ are the cardinality operator, the Sigmoid function, and the output of the channel $m$ (boundary or center) of the loss prediction module, respectively.\\
 Now, we have everything for the loss of training phase II in place:
\begin{equation}
\label{eq:phaseII}
    \mathcal{L}_{\textit{II}}=\lambda_{0}\mathcal{L}^{c}_{\textit{L}} +\lambda_{1}\mathcal{L}^{b}_{\textit{L}}+\sum_{k=1}^{3}\lambda_{k+1}\mathcal{L}^{k}_{\textit{seg}}
\end{equation}
Where $\lambda_{k}$'s are the regularization constants chosen by cross-validation and parameter search.
\vspace{-25pt}
%-------------------------------------------------------------------------
\vspace{-3pt}
\subsection{Selection Strategy}
We follow a hierarchical approach to breakdown the distribution of the data through which we measure the uncertainty of samples by starting to look from a low field of view (pixel level:\ top level of the hierarchy) going incrementally to the highest field of view (Image level).\ At each field of view,\ we assess the uncertainty level within different scopes to find regions needing attention for sampling.\\
\indent\textbf{Pixel Level-Getting Aware of the Maturity}: We start the uncertainty assessment by analyzing the pixels individually.\ At this level, we feed all the samples to the trained model and for the pixels in the unlabeled pool, measure the uncertainty based on: 
\begin{multline}
\label{eq:pixel_level}
   u(x)=\Bigl( H(\hat{Y}_{x})+W^{1}_{x}JS(\hat{Y}_{x},\hat{Y}^{1}_{x})+W^{2}_{x}JS(\hat{Y}_{x},\hat{Y}^{2}_{x})+W^{3}_{x}JS(\hat{Y}_{x},\hat{Y}^{3}_{x}) \Bigr)\\\Bigl[\Bigl(1-\delta(x)\Bigr)e^{\sigma(\hat{Y}_{L-b}^{x})}+\delta(x)e^{\sigma(\hat{Y}_{L-c}^{x})}\Bigr] 
\end{multline}
\begin{equation*}
\delta(x)= 
\begin{cases}
    1,& \text{if }x\in center \\
    0, & \text{otherwise}
\end{cases}
\end{equation*}
Here $H(P)$, $JS(P_{1},P_{2})$, $W^{k}_{x}$, and $\hat{Y}^{k}_{x}$ denote entropy of probability distribution $P$, Jenson-Shannon  divergence between distributions $P_{1}$ and $P_{2}$, weight map predicted by the weight prediction block for the $k$-th head, and the output distribution by the $k$-th head for pixel $x$, respectively. Via Eq. \ref{eq:pixel_level}, we measure the uncertainty of the model for pixel $x$ by checking: i) the entropy of the final output distribution (reflected by the first term), ii) the similarity of the final output distribution to those of the varied-maturity heads (reflected by the second-fourth terms). The intuition is that the easier a pixel is to classify, the less depth is needed to produce an output similar to the final one. The importance of each term is determined by the weight map corresponding to the pixel and segmentation head ($W^{k}_{x}$). Lastly, iii) error-proneness of the pixel (reflected by the exponential terms), determined based on the score given for the pixel by its corresponding channel (center or boundary) of the loss prediction module.\\
\indent\textbf{Superpixel Level}: Next, we zoom out and look through the superpixel level. By definition, a superpixel is a group of perceptually similar pixels. First, we assign each superpixel to its dominant label $Do(s)$ (the predicted class for the majority of the pixels within that superpixel \cite{SAL}); however, unlike \cite{SAL}, we estimate the probability of the class $C_{i}$ within cluster $k$ ($cl_{k}$) by counting the superpixels with the dominant label of $C_{i}$:
\begin{equation}
    P_{cl_{k}}(C_{i})=\frac{|\{s:Do_{s}=C_{i}\&s\in cl_{k}  \}|}{|\{ s:s\in cl_{k}\}|}
\end{equation}
Now, we assess the uncertainty of each superpixel by:
\begin{equation}
\label{eq:superpixel}
u(s)=\frac{\underset{x \in s}{\operatorname{\sum}} u(x)}{|\{ x|x\in s\}| }e^{-P_{cl_{k}}(Do(s))} \textit{  s.t.  } s \in cl_{k} 
\end{equation}
Based on Eq. \ref{eq:superpixel} the uncertainty of a superpixel in a cluster is proportional to the average uncertainty of its pixels and inversely proportional to the abundance of its dominant label.\\
\indent\textbf{Cluster Level}: Having the uncertainty of the superpixels in each cluster, we now assess the uncertainty of each cluster:
\begin{equation}
\label{eq:cluster}
    u(cl_{k})=\frac{\sum _{s\in cl_{k}}u(s)}{|\{ s:s\in cl_{k}\}|}
\end{equation}
The uncertainty of each cluster determines the budget it is assigned in the sample selection step. The more uncertain a cluster is, the larger budget it is assigned to:
\begin{equation}
\label{eq:budget}
    B_{cl_{k}}=\ceil*{\frac{u(cl_{k})}{\sum_{j=1}^{N_{clusters}}u(cl_{j})}B_{t}}
\end{equation}
Where $B_{cl_{k}}$, $B_{t}$, and $\ceil*{.}$ are the budget assigned to cluster $k$, the total budget, and the ceil function, respectively.\\ 
\indent\textbf{Image Level}: Once the uncertainty scores of all the lower fields of view are figured out, we query pixels for each image based on: 1) considering budget dedicated to each cluster in the image, 2) finding superpixels with highest uncertainty within that cluster, and 3) selecting pixels with highest uncertainty within these superpixels. It is worth mentioning that traversing from the top to the bottom of the hierarchy helps us achieve a global insight of the uncertainty across different regions of the image, while the trip back to the top aids with finely detecting and selecting a small, yet impactful number of samples for annotation. 
\vspace{-3ex}
\begin{figure*}
    \centering
    \includegraphics[width=1\textwidth]{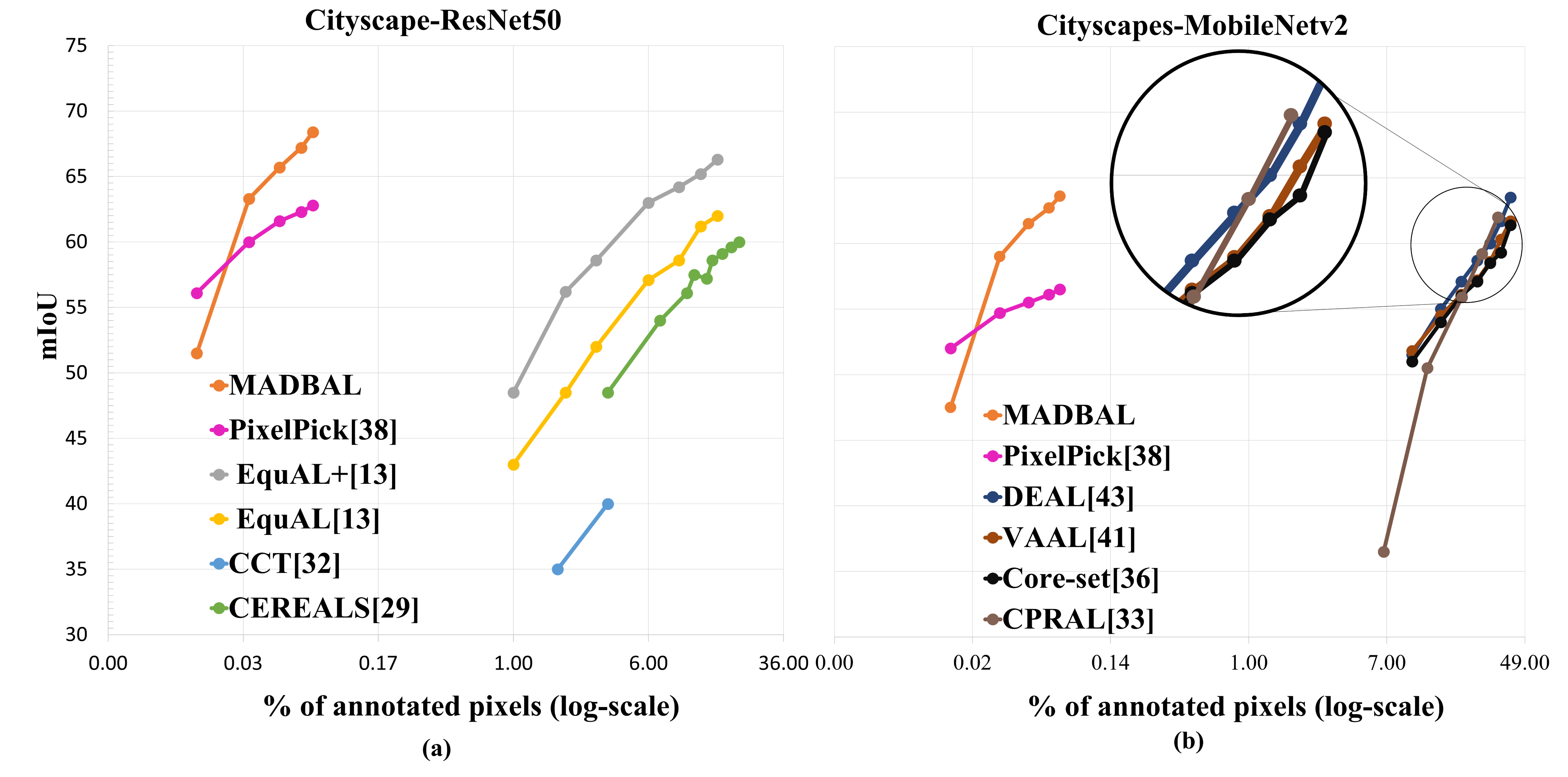}
    \vspace{-25pt}
    \caption{Comparison with SOTA on Cityscapes with two of the most popular backbones in AL methods (ResNet50 and MobileNetv2).}
    \label{fig:cityscapes_absol}
    \vspace{-15pt}
\end{figure*}
\section{Experiments \protect\footnote{Find the numerical data for the plots and codes here:\href{https://github.com/yazdaniamir38/MADBAL}{\underline{github/MADBAL}}}}
\noindent\textbf{\textbf{Implementation Details}}: We evaluate MADBAL on Cityscapes \cite{cityscapes} and Pascal VOC 2012 \cite{VOC} datasets by training on samples from the training set and testing on the validation set. Our initial labeled pools have 10 and 20 randomly selected pixels per image for VOC and Cityscapes, respectively. The AL budget in our experiments is 10 and 20 pixels per image for VOC and Cityscapes, respectively. We use SEEDS algorithm \cite{seeds} for superpixel extraction and set the number of superpixels per images for both datasets equal to the number of squares when the image is divided to the squares of size $16\times16$. we cluster these superpixels; however, superpixels are of irregular (not necessarily rectangular or vector) shapes, which is not acceptable by K-means. To address this, we first fit the superpixel at the center of a rectangular patch with minimum size and then resize that patch to a certain size ($16\times16$). Consequently, we feed the resized patch to the backbone of a pretrained VGG 16 \cite{vgg16} and apply K-means to the extracted feature vectors. This way the clustering would be done based on the perceptual properties of the superpixels. For each dataset, we conduct our experiments three times with its most prevalent backbones in the literature: ResNet50 \cite{resnet50}, MobileNetv2 \cite{mobilenetv2}, and MobileNetv3 \cite{mobilenetv3} for Cityscapes and ResNet50 and MobileNetv3 for VOC.

During training, for Cityscapes, we acquire random crops of size $768\times768$ from the samples and for VOC random crops of size $256\times256$. Our models are deployed using Pytorch and we use stochastic gradient descent optimizer with an initial learning rate of 0.01, momentum of 0.9 and a poly learning rate scheduler decaying the learning rate from the initial value to zero linearly through 150 epochs (for phase I) and 30 epochs (for phase II). The hyperparameres in Eq. \ref{eq:phaseII} are 1, 1, 0.05, 0.1, and 0.15 respectively, selected via cross-validation on initial labeled pools of Cityscapes.

\begin{figure*}
    \centering
    \includegraphics[width=1\textwidth]{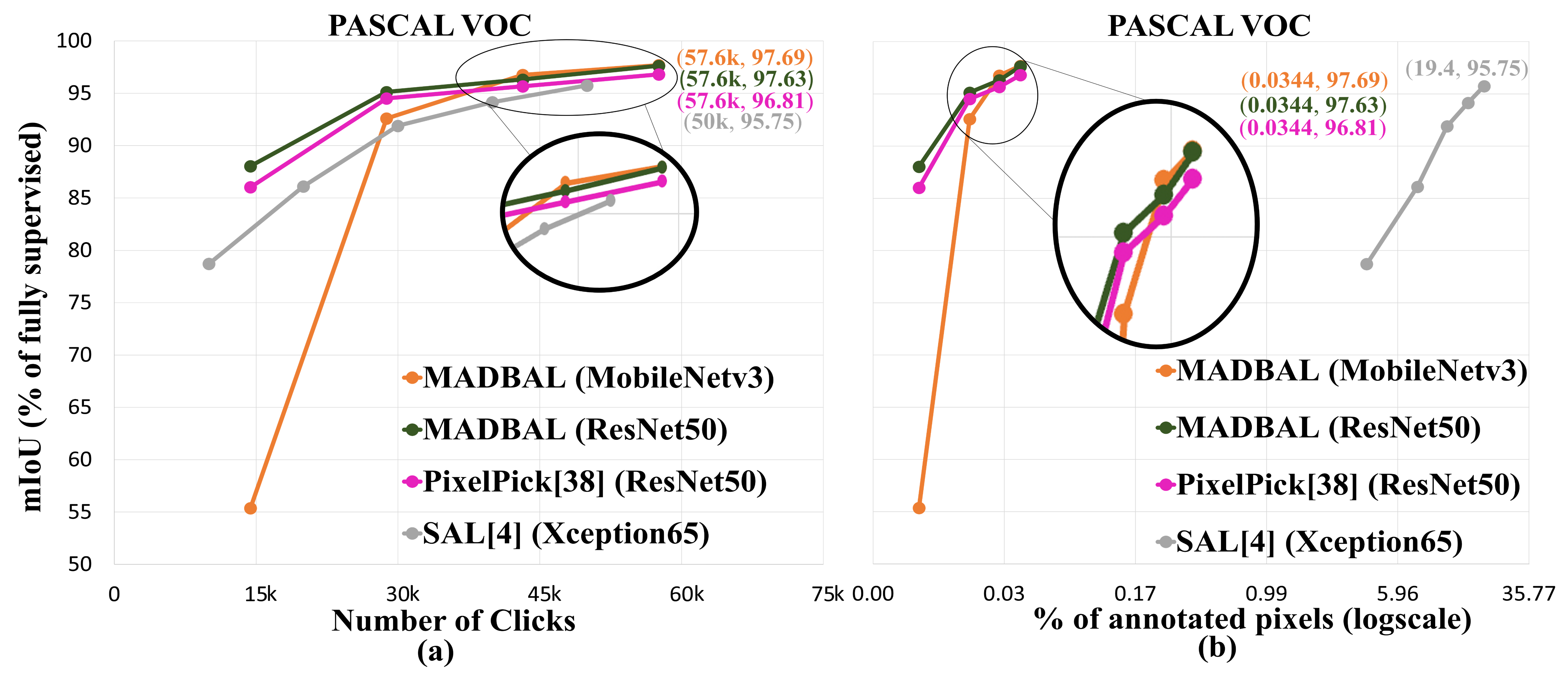}
    \vspace{-25pt}
    \caption{Comparison with SOTA on VOC. w.r.t. number of clicks \textbf{(a)} and percentage of annotated pixels \textbf{(b)}.  }
    \label{fig:voc_absol}
    % \vspace{-15pt}
\end{figure*}

\begin{figure*}
    \centering
    \includegraphics[width=1 \textwidth]{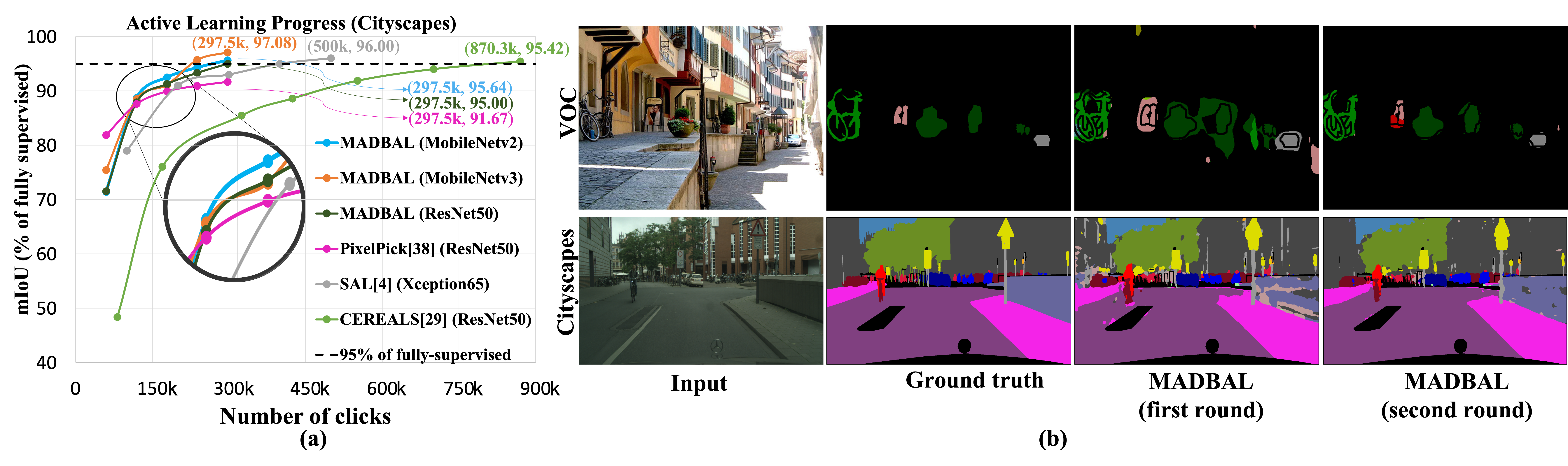}
    \vspace{-25pt}
    \caption{\textbf{(a)}: Performance results on Cityscapes based on the number of clicks (for each method the exact values of the last AL step are shown with its matching color). \textbf{(b)}: Visualization results on both datasets. First round of AL is completed with 20 and 10 pixels per image, and second round with 40 and 20 for Cityscapes and VOC, respectively.}
    \label{fig:cityscapes_click}
\end{figure*}

\noindent\textbf{Comparison with State of the Art}: Figs. \ref{fig:cityscapes_absol} , \ref{fig:voc_absol}, and \ref{fig:cityscapes_click}a report the average \emph{mean intersection over union} (mIoU) of three repetitions of our experiments for each dataset and backbone w.r.t. different budget measures. We can observe that our annotation cost is two orders of magnitude lower than the majority of the SOTA (regardless of the backbone) w.r.t. the percentage of annotated pixels, while outperforming SOTA with a significant margin w.r.t. number of clicks. Moreover, MADBAL starts with a lower performance than Shin \emph{et al.}'s \cite{pixelpick}, which is mostly depending on the richness of the initial labeled pool, and makes considerably large leaps and outperforms their method quickly. This implies  MADBAL's effectiveness in selecting the most important samples early on. To get a qualitative sense of these leaps, Fig. \ref{fig:cityscapes_click}b visualizes the performance of MADBAL through the first two AL steps on a validation sample from each dataset. Finally, Tab. \ref{tab:VOC} compares various weakly-supervised and interactive weak supervision methods on VOC, confirming the benefits of MADBAL trained with only 20 pixels per image. 
% Finally, following a similar approach to \cite{SAL} has in reporting the results, we compare the performance of MADBAL with respect to the budget needed to acquire 95$\%$ of the fully supervised performance of the corresponding backbone. Tab. S1 showcases this for different models. The numbers represent the percentage of number of clicks needed for each method normalized to the total number of clicks needed for annotatinng the Cityscapes training set.

\begin{table}[!ht]
    \centering
        \caption{Comparison with Weakly-supervised methods and PixelPick \cite{pixelpick} on VOC.}
    \begin{tabular}{|c|c|c|c|}
    \hline
         Method&Backbone&Train set (anno. type)&mIoU  \\
         \hline
         \multicolumn{4}{|l|}{\textbf{Weakly-supervised methods}}\\
         GAIN \cite{GAIN}&VGG16&10.k imgs (classes)&55.3\\
         MDC \cite{MDC}&VGG16&10.k imgs (classes)&60.4\\
         DSRG \cite{DSRG}&ResNet101&10.5k imgs (classes)& 61.4\\
         FickleNet \cite{FickleNet}&ResNet101&10.5k imgs (classes)& 64.9\\
         BoxSup \cite{BoxSup}&VGG16&10.5k imgs (boxes)& 62.0\\
         ScribbleSup \cite{ScribbleSup}&VGG16&10.5k imgs (scribbles)&63.1\\
         \hline
          \multicolumn{4}{|l|}{\textbf{Interactive weak supervision}}\\
          PixelPick \cite{pixelpick}&ResNet50&1.5k imgs (20 pixels per image)&65.6\\
          MADBAL&ResNet50&1.5k imgs (20 pixels per image)&\textbf{72.4}\\
          \hline
    \end{tabular}
    \label{tab:VOC}
    \vspace{-10pt}
\end{table}
\noindent\textbf{Ablation Study}: We validate our design of MADBAL by conducting experiments devised to show how presence and absence of various components affect the performance. For these experiments, we incorporate MobileNetV3 \cite{mobilenetv3} backbone and Cityscapes \cite{cityscapes} dataset.
% \begin{table}[]
%     \centering
%      \caption{Comparison between different methods in terms of the annotation budget of Cityscapes training set needed for achieving 95$\%$ of their fully supervised backbone (the lower the better).}
%      \resizebox{1\textwidth}{!}{
%     \begin{tabular}{|c|c|c|c|c|}
%     \hline
%          Method&SAL\cite{SAL}&CEREALS\cite{cereals}&Metabox+\cite{metabox}&Entropybox+\cite{metabox}\\
%          \hline
%          Budget&7.85$\%$&35.29$\%$&10.47$\%$&10.25$\%$\\
%          \hline
%          Method&\multicolumn{2}{|c|}{MADBAL (MobileNetv2)}& MADBAL (MobileNetv3)& MADBAL (ResNet50)\\
%          \hline
%          Budget&\multicolumn{2}{|c|}{\textbf{5.28}$\%$}&\textbf{4.49}$\%$&\textbf{5.84}$\%$\\
%          \hline
%         %  MADBAL (MobileNetv3)&MADBAL (MobileNetV2)&MADBAL (ResNet50)
%     \end{tabular}}
%     \label{tab:cityscapes_clicks}
% \end{table}
\textbf{i) Effect of maturity-awareness and loss prediction}
in this set of experiments is focused through four different AL scenarios: \textbf{1)} AL with MADBAL. \textbf{2)} AL with a modified MADBAL in which weight prediction block is dropped and averaging is used instead (i.e. $\frac{1}{3}$ is used instead of weight maps). This is to show the essence of giving different importance to different intermediate features and its benefits for recognition of impactful pixels (denoted with ``Averaging''). \textbf{3)} AL with a modified MADBAL in which the loss prediction module only accesses the backbone features (i.e. no inputs from the varied-maturity heads) and the uncertainty score calculation (Eq. \ref{eq:pixel_level}) does not have Jenson-Shannon divergence terms. This helps observe the effect of maturity-awareness directly by removing the corresponding terms in uncertainty score formulation (denoted with ``No maturity-awareness''). \textbf{4)} AL with vanilla backbone (no loss prediction module, weight prediction block, and varied-maturity heads) to analyze the performance of MADBAL solely relying on distribution breakdown (denoted with ``Vanilla'').  

\textbf{ii) Effect of distribution breakdown}, as another important piece of novelty during the sample selection stage, is studied via 3 AL scenarios: \textbf{1)} AL with MADBAL to show the benefits of the distribution breakdown, \textbf{2)} AL with modified MADBAL in which superpixels are assigned to random clusters while keeping the number of clusters the same. This is to observe how clustering superpixels based on their perception plays a role in detecting the most uncertain samples while keeping the diversity (denoted with ``Random dist-breakdown''). \textbf{3)} AL with modified MADBAL which does not benefit from distribution breakdown at all. In this scenario N pixels with highest uncertainty scores in each image are queried for annotation (denoted with ``No dist-breakdown'').

\begin{figure*}
    \centering
    \includegraphics[width=.9 \textwidth]{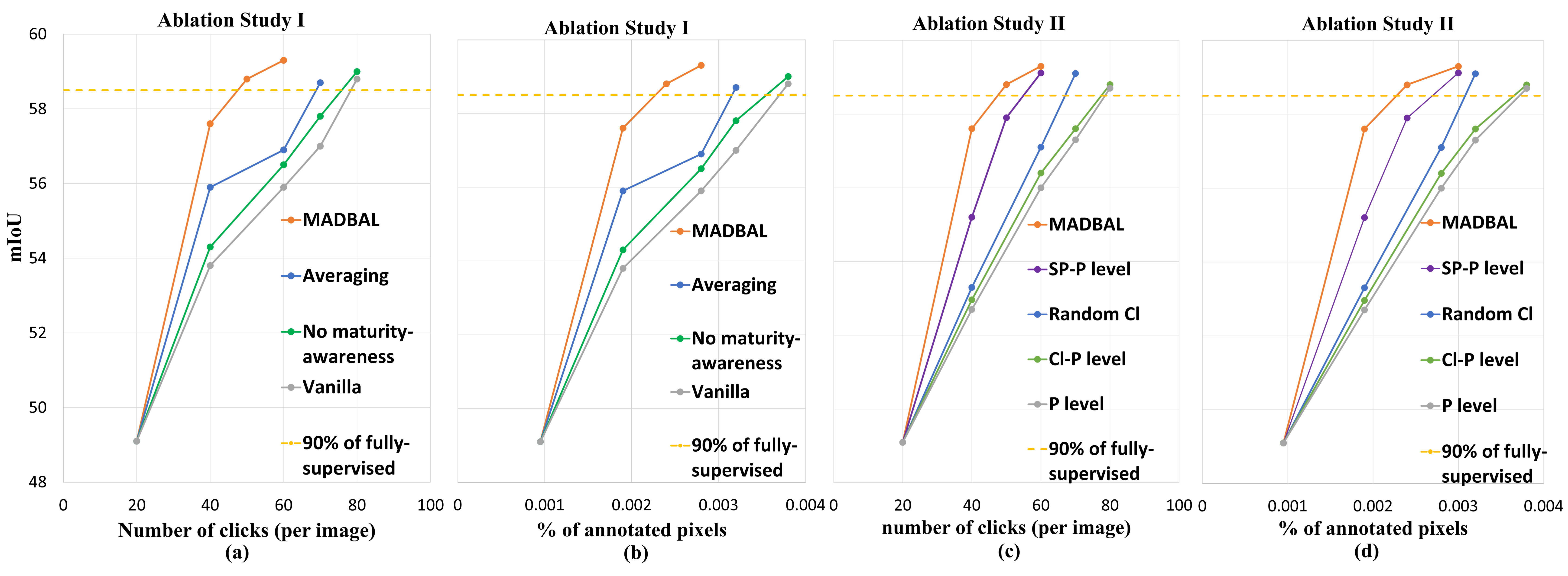}
    \vspace{-10pt}
    \caption{Ablation study on the effect of proposed components. (\textbf{a}), (\textbf{b}): the effect of maturity-awareness and loss prediction. (\textbf{c}), (\textbf{d}): the effect of distribution breakdown. When all the components are put into work the highest annotation efficiency is achieved. The more components are dropped, the more degradation on the efficiency is resulted.  }
    \label{fig:ablation_1}
\vspace{-12pt}
\end{figure*}

We continue each AL progress until 90$\%$ performance of fully-supervised model ($0.9\times68.5=65.1\%$) is achieved. Fig. \ref{fig:ablation_1} depicts the results. As expected, MADBAL, with its fully extended features, achieves $90\%$ performance with only 50 pixels per image owing to all the devised components. For the first ablation study, the second-best performance belongs to ``Averaging,'' which matches the intuition as the algorithm benefits from the varied-maturity heads both for loss prediction and uncertainty score calculation; however, removing learnable weight maps adversely affects its performance compared to MADBAL. ``Vanilla'' and ``No maturity-awareness'' show the worst performances due to missing the critical components. Between the two, ``Vanilla'' is inferior as it does not benefit from the maturity-awareness nor from the loss prediction module. ``No maturity-awareness,'' on the other hand, shows better performance owing to loss prediction module helping with better assessment of samples' uncertainty. For the second ablation, it is worth noticing that ``Random dist-breakdown'' is still showing a better performance compared to ``No dist-breakdown.'' This can be attributed to the inevitable diversity the clustering (whether it be a perception-based clustering algorithm or random clustering) brings to the pulled samples in each round of sample selection. In other words, by grouping the superpixels, each of which corresponds roughly to an object class, we prevent ``over-selection'' of pixels belonging to the same object category in each step. Hence, despite its lower performance, ``Random dist-breakdown'' is still able to achieve higher performance than ``No dist-breakdown.''

\section{Conclusion}
In this work we proposed an active learning framework for semantic segmentation by integrating maturity-awareness and distribution breakdown. Maturity-awareness helps develop an effective understanding and recognition of the most critical pixels for performance improvement, while distribution breakdown provides a hierarchical approach to have an inclusive insight of the data distribution across different fields of view. Combined with a novel uncertainty formulation, the proposed MADBAL is shown to outperform many state of the art methods with significant margin. MADABAL can significantly reduce training burdens and also be impactful for tasks where annotation is expensive and not readily available. 
%------------------------------------------------------------------------- 
\bibliography{egbib}
\end{document}